# Deep Learning based Tomato Disease Detection and Remedy Suggestions using Mobile Application

Yagya Raj Pandeya[1, 2*], Samin Karki[1], Ishan Dangol[1], Nitesh Rajbanshi[1]

[1]Department of Computer Science and Engineering, School of Engineering, Kathmandu University, Nepal
[2]Guru Technology Pvt.Ltd., Kathmandu, Nepal

{yagyapandeya, samin.karki, ishandongol, niteshrajbanshi}@gmail.com

*Abstract*—**We have developed a comprehensive computer system to assist farmers who practice traditional farming methods and have limited access to agricultural experts for addressing crop diseases. Our system utilizes artificial intelligence (AI) to identify and provide remedies for vegetable diseases. To ensure ease of use, we have created a mobile application that offers a user-friendly interface, allowing farmers to inquire about vegetable diseases and receive suitable solutions in their local language. The developed system can be utilized by any farmer with a basic understanding of a smartphone. Specifically, we have designed an AI-enabled mobile application for identifying and suggesting remedies for vegetable diseases, focusing on tomato diseases to benefit the local farming community in Nepal. Our system employs state-of-the-art object detection methodology, namely You Only Look Once (YOLO), to detect tomato diseases. The detected information is then relayed to the mobile application, which provides remedy suggestions guided by domain experts. In order to train our system effectively, we curated a dataset consisting of ten classes of tomato diseases. We utilized various data augmentation methods to address overfitting and trained a YOLOv5 object detector. The proposed method achieved a mean average precision of 0.76 and offers an efficient mobile interface for interacting with the AI system. While our system is currently in the development phase, we are actively working towards enhancing its robustness and real-time usability by accumulating more training samples.**

## I. Introduction

In many developing countries, numerous farmers lack direct access to agriculture experts for consultation regarding their crop issues and remedies. These farmers typically rely on traditional farming methods passed down through generations. However, due to climate change resulting from global warming, the general crop calendar has been significantly affected on a global scale. Furthermore, new diseases are emerging in crops. Therefore, the presence of domain experts is crucial for local farmers within communities who adhere to traditional farming practices and have low annual incomes. These farmers heavily depend on their crops as their primary source of income, with limited alternatives for additional earnings. Unfortunately, these farmers often face challenges in hiring domain experts for consultations, as they have limited or no access to direct or indirect communication channels supported by the government. Consequently, many farmers in Nepal and India resort to desperate measures, including suicide, when their crops are devastated by diseases.

The objective of this project is to develop an AI-guided system for vegetable disease identification and remedy suggestions, tailored to the understanding of local farmers in Nepal. Initially, we trained and evaluated our object detection model using a two-stage detector, which yielded high evaluation scores. However, the computational complexity and time delay associated with the two-stage detector made it unsuitable for real-time object detection. To address this issue, we deployed our model on lightweight devices such as smartphones. We trained our tomato dataset using a one-stage detector model called YOLOv5 (You Only Look Once, version 5). This model represents the state-of-the-art in object detection and demonstrates better performance than the two-stage detector in many cases.

Numerous studies have been conducted on disease classification in plants and crops [1, 2], and some research [3, 4] has focused on vegetable disease localization using deep learning technology. The Plant Village dataset [5], comprising 54,305 healthy and infected plant leaves, is a well-known dataset for plant disease classification. Specific research efforts have also been directed towards vegetable diseases. For instance, Rashid et al. [6] proposed a classification system for potato diseases, while other studies [7, 8, 9] have explored the classification and localization of tomato diseases and pests. Furthermore, recent research has extended the idea of disease detection and segmentation to strawberries [10].In contrast to previous research, our focus lies in addressing the issues specific to our community, which arise due to changes in pets (organisms that harm crops) resulting from climate change and seasonal variations. To develop an efficient system that accurately understands vegetable diseases and pets present in Panchkhal. Our ultimate goal is to find the best solution for recognizing vegetable diseases, localizing them, and providing optimal remedies under the supervision of domain experts.

We initiated our work by focusing on tomato diseases, as they are a major concern for vegetable crops in Nepal. The disease patterns and remedies for tomato diseases are often similar to those of other crops such as potatoes, chili, capsicum, and more. To cater to the needs of local farmers, we developed a user-friendly smartphone interface that allows them to easily identify their vegetable diseases and understand the corresponding remedy procedures in Nepali language. Our system only requires a smartphone to access expert assistance. In contrast to previous studies on vegetable disease identification, we developed an automated system that can identify local diseases and created a production pipeline to

This work was funded by Kathmandu University, Community Engagement Division, Office of Vice-Chancellor.
* Corresponding author

efficiently assist local farmers. Previous projects mainly focused on developing neural networks for disease identification but did not integrate them into real-time systems to harness their full potential. In our case, we deployed our trained model on a mobile application that is freely accessible to local farmers. The key contributions of this work are as follows:

A. *Addressing community problems:* The main objective of our work is to address the detection and remedy suggestions for vegetable diseases, fulfilling the specific needs of local farmers using local data resources. We obtained valuable feedback through formal and informal interactions with local farmers and elected representatives of the local government to better understand their requirements.

B. *Nepali language interface:* In response to suggestions from local farmers and government representatives, we developed a fully functional system that operates in Nepali language, ensuring clear comprehension for local farmers when dealing with their vegetable disease-related issues. The detected vegetable diseases, symptoms, and remedies are presented in Nepali language for enhanced understanding.

C. *Lightweight and powerful:* Our computer system is designed to be lightweight and compatible with smart mobile devices that possess limited computational resources. Despite its lightweight nature, the system achieves high precision in vegetable disease detection.

D. *Real-time system:* Our system is implemented as a real-time mobile application. It can also operate in offline mode, updating its features only when it is connected online.

E. *Lifelong learning system:* Our system has the capability to self-correct errors and adapt to changes in diseases over time.

F. *User-friendly:* We have designed the system in an abstract manner to provide users with an easy interface and a comfortable working environment.

The remaining sections of this report include details about the tomato disease dataset, neural network architecture methods, operational workflow of the mobile application, results and discussions, and finally, a summary of our project work.

## II. TOMATA DISEASE DATASET

The deep learning-based disease recognition system needs more data to train them well. As a result, our project began with the task of collecting representative sample data that encompassed the various vegetable diseases prevalent in the local area. Ten colleagues actively participated in the data collection process, which involved annotation and information retrieval from the gathered samples. The data was collected from both local traditional open farms and commercially protected farms, with farmers being interviewed using comprehensive questionnaires. We also sought domain expertise from agricultural experts to deepen our understanding of vegetable diseases. As of the time of writing, we have accumulated a dataset consisting of ten tomato diseases, including a background class. Figure 1 illustrates the disease names along with their corresponding samples.

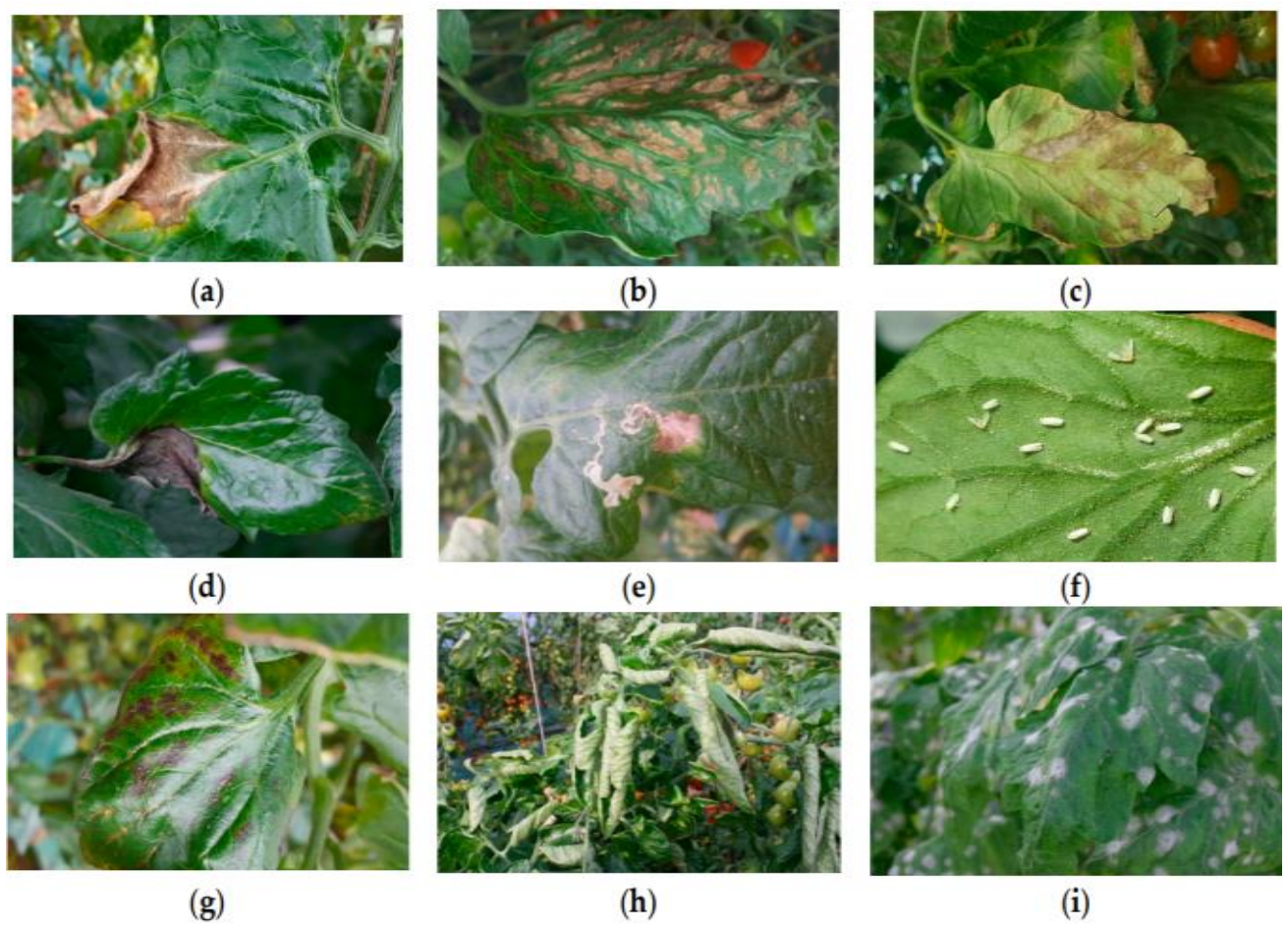

Fig. 1. The sample data from each class of tomato disease. (a) Gray mold, (b) Canker, (c) Leaf mold, (d) Plague, (e) Leaf miner, (f) Whitefly, (g) Low temperature, (h) Nutritional excess or deficiency, and (i) Powdery mildew.

To overcome the scarcity of data, we dedicated one month to the collection of tomato disease data. We focused on providing detailed descriptions of the diseases and the corresponding remedy information for the most common ailments. The data collection process encompassed various stages of plant growth, and we categorized the diseases into nine distinct classes. In total, we gathered approximately 4000 image samples from local farms and commercially protected farms. However, it is important to note that the dataset we collected is imbalanced, meaning that there is a variation in the number of data samples for each disease class. The distribution of the dataset is displayed in Figure 2, highlighting the imbalance within the dataset. For instance, the "Leaf mold" disease, or "lmold," has around 7000 samples, while some other disease samples number less than 100.

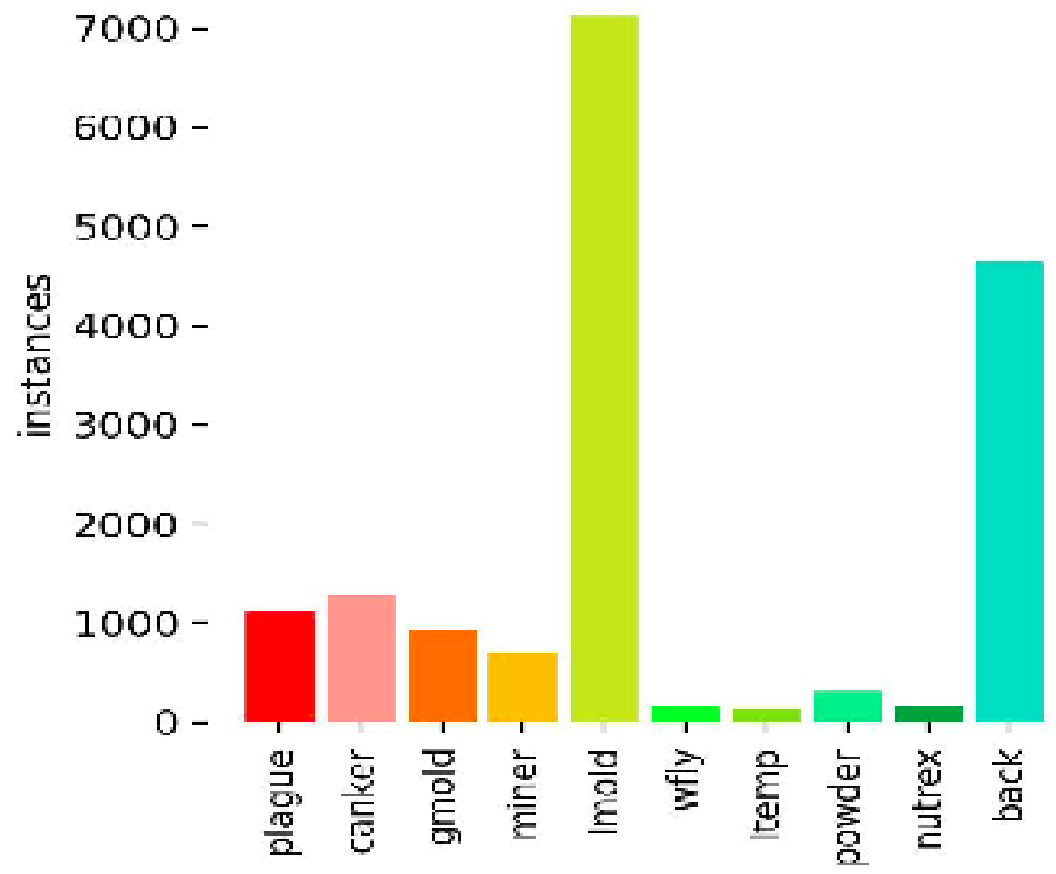

Fig. 2. The objects distribution in tomato disease detection dataset

After collecting the data samples from local farms, we proceeded with the crucial step of annotating the data for tomato disease detection. Data annotation involves properly labeling the data to facilitate effective learning within our neural network-based system. To address potential overfitting issues in the neural network, we employed several augmentation techniques. These techniques included mosaic augmentation, scaling, mixup, translation, rotation, and PCA color augmentation. By utilizing these augmentation methods, we aimed to diversify and enhance the dataset. Throughout the project, we maintained regular discussions with vegetable experts to gain valuable insights into disease types, characteristics, solutions, and potential future treatments. Based on the recommendations from these experts, we labeled the tomato diseases and compiled a comprehensive list of disease symptoms, characteristics, remedies, and cures, as outlined in Table I. For better understanding, we have provided examples of data labeling for object detection in Figure 3.

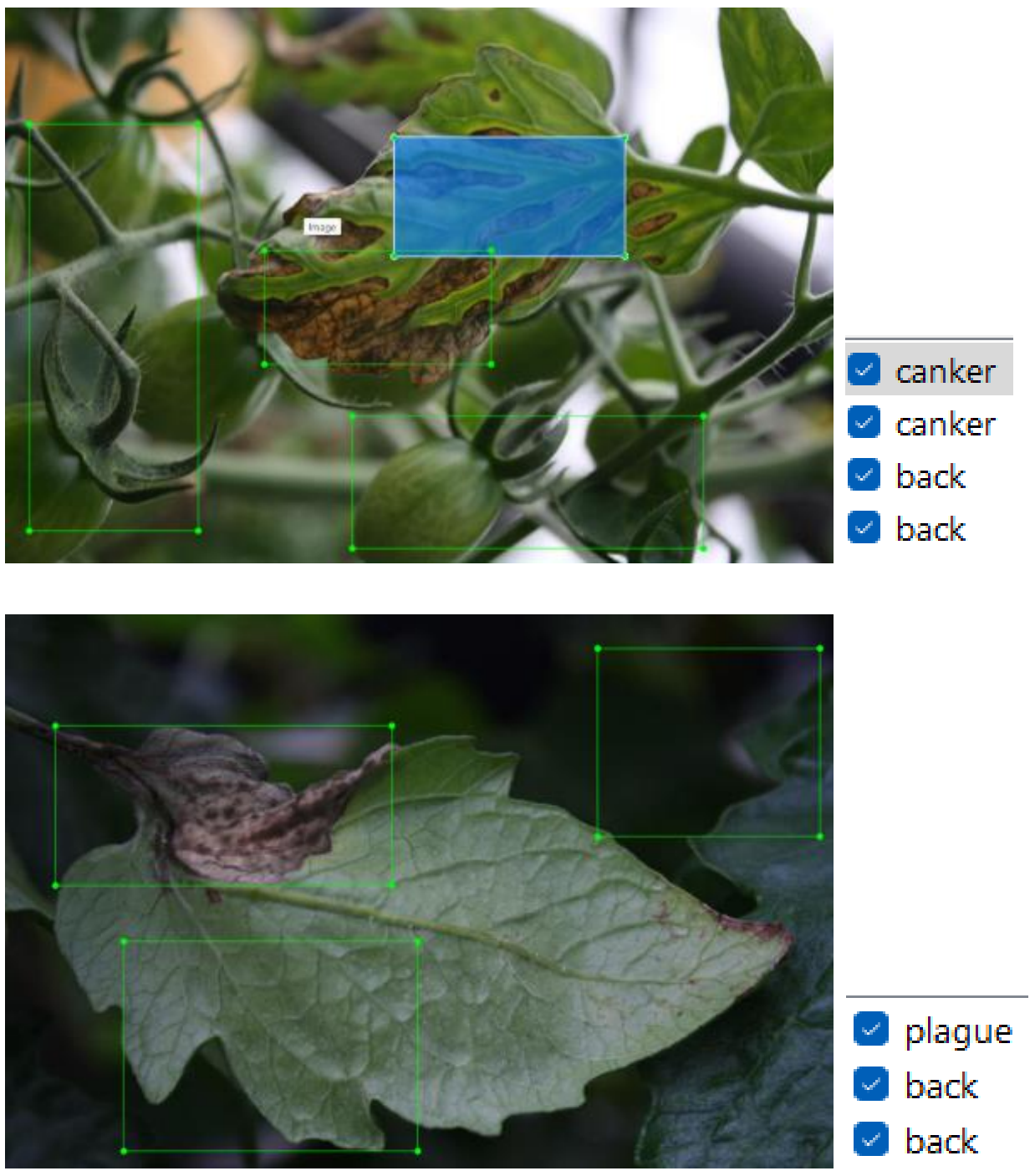

Fig. 3. Some data samples of tomato disease annotation and corresponding labels

## III. Methods

The research work commenced with the first phase, which involved the collection of vegetable disease data in collaboration with agricultural scientists. This data was then utilized to train an artificial intelligence-based deep learning system, allowing it to automatically learn and analyze various input data. The system's intelligence and capabilities are directly proportional to the diversity and range of data it is exposed to. To facilitate this process, an Android-based application was developed, enabling farmers to capture images of infected vegetables. These captured images are then transmitted to a server for further processing. The server analyzes the data and extracts the relevant information, which is subsequently sent back to the farmer. The farmer can conveniently view the obtained useful information on their screen. The basic workflow of this method is illustrated in Figure 4.

TABLE I. Description of a particular disease "Powdery Mildew" of tomato for mobile application

| | |
|---|---|
| Disease class (रोग किरा ) | सेतो दुसी रोग वा खरानी रोग (Powdery Mildew) |
| Symptoms (लक्षणहरू) | • सुरूमा पातको माथिल्लो सतहमा हल्का सेतो वा कैला धब्बाहरू देखा पर्दछन् ।<br>• यो रोगको आक्रमण बढ्दै जाँदा पात पुरै सेतो देखिने र उक्त सेतो पाउडर पात, डाँठ र फलमा पनि देखा पर्दछ ।<br>• अन्तमा पुरै बोट पहेंलिने, घुम्रीने र सुक्न थाल्दछ ।<br>• बिरुवा परिपक्क हुन नपाँउदै पातहरू सुकेर मर्दछन् । जसले गर्दा फलहरू साना, चाउरी परेका र ढिलो पाक्न गई फल कम गुणस्तरका हुन्छन् । |
| Prevention (रोकथाम) | • रोग लागेका पात, डाँठ र हाँगाहरू कटिङ्ग गरी टाढा लगेर नष्ट गर्ने । कटिङ्ग गरेर आएका पात, डाँठलाई जथाभाबी छाड्नाले रोगको प्रकोप बढ्न जान्छ ।<br>• बिरुवा रोप्नु अगाडी नै राम्रोसँग सरसफाई गर्ने । पुरानो बोटलाई उखली राम्रोसँग नष्ट गर्ने ।<br>• माटोमा झरेका पुराना पात वा फलहरूलाई संकलन गरी नष्ट गर्ने । बिरूवा पातलो लगाउने ।<br>• पानीको निकासको राम्रो प्रबन्ध मिलाउने ।<br>• माथिबाट सिंचाई नगर्ने । हावाको राम्रो चलखेल गराउने त्यसको लागि समय समयमा पात काँटछाँट गरि राख्ने ।<br>• रोग अवरोधक जात लगाउने । |
| Remedy (उपचार) | • प्राङ्गारिक खेतीमा हो भने शुरू देखी नै खाने सोडा (Baking soda) १० ग्राम प्रति लिटर पानीमा मिसाएर १० दिनको फरकमा २-३ पटक स्प्रे गर्ने ।<br>• बेसार ३ ग्राम र सेतो खरानी ३ ग्राम प्रति लिटर पानीमा राम्रोसँग मिसाउने र पतलो कपडाले छानेर मात्र पम्पमा मिसाउने र बिहान वा बेलुकी पख ७-१० दिनको फरकमा स्प्रे गर्ने ।<br>• रासयनिक तरीकामा तलको मध्य कुनै प्रयोग गर्न सकिन्छ ।<br>• फाटफुट मात्रै रोग देखेको अवस्थामा नाईटर Nitor (Thiophanate Methyl 70%WP) = १.५ ग्राम स्टीकअन = ०.५ एम. एल. प्रति लिटर पानीमा मिसाई ७ देखि १० दिनको फरकमा २ देखि ३ पटक स्प्रे गर्ने ।<br>• क्यारेथेन Karathine (Dinocarp 48%) = १-१.५ एम.एल. स्टीकर = ०.५ एम. एल. प्रति लिटर पानीमा मिसाई ७ - १० दिनको फरकमा २ देखि ३ पटक चोटि स्प्रे गर्ने ।<br>• सल्फर Sulphur 80% WP = १.५-२ ग्राम + स्टीकर एम.एल. प्रति लिटर पानीमा मिसाई ७ देखि १० दिनको फरकमा २ देखि ३ पटक साँझ वा बिहानी पख स्प्रे गर्ने प्लाष्टिक घर मा बोट सानो अवस्थामा मात्रै प्रयोग गर्ने ।<br>• ईस्तार टप STAR TOP (Azoxystrobin 20%+Difenconizole 12.5%) = १ एम.एल. + स्टीकर = ०.५ एम.एल. प्रति लिटर पानीमा मिसाई ७ देखि १० दिनको फरकमा २ देखि ३ पटक स्प्रे गर्ने । |

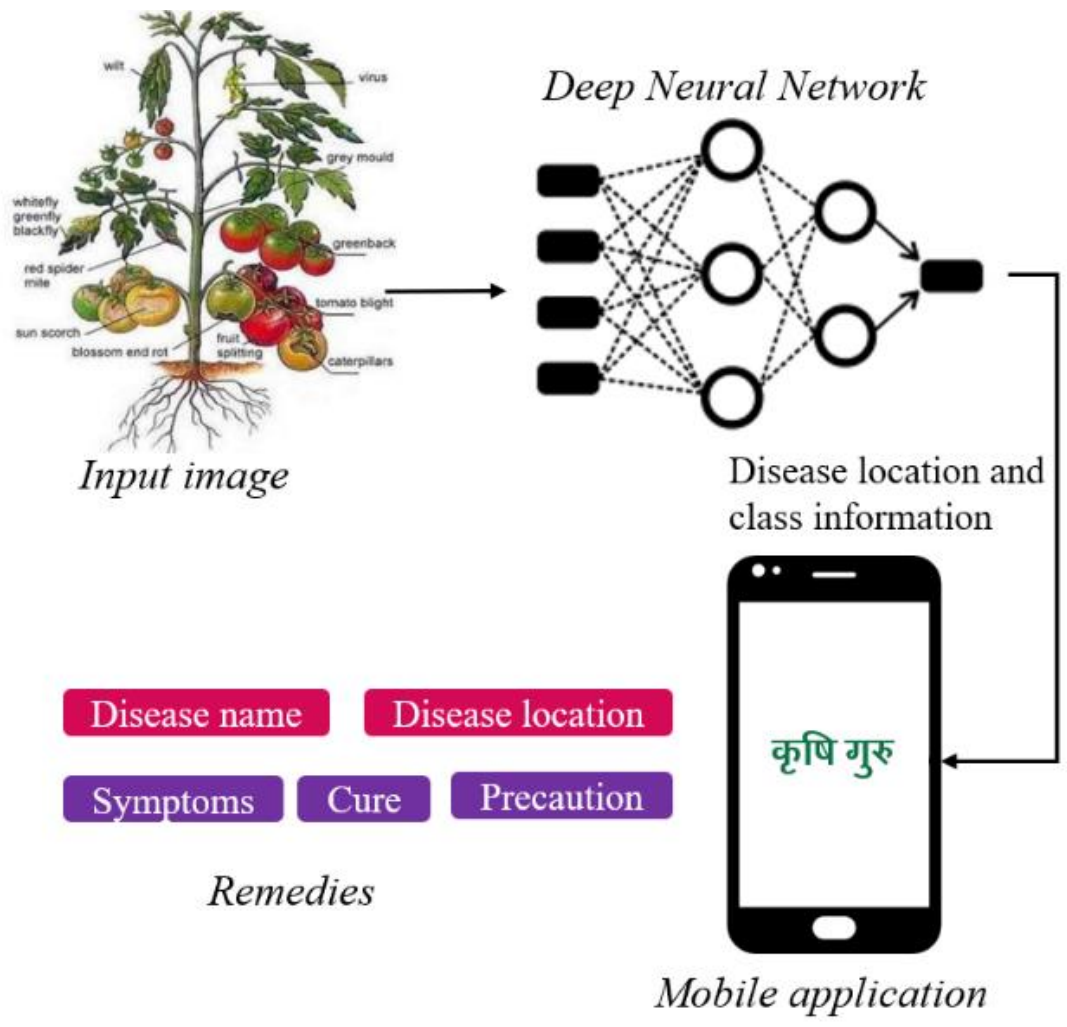

Fig. 4. Basic workflow of proposed system

## A. *Deep neural network*

Deep neural networks are employed to accurately detect vegetable diseases and assign them appropriate labels. The system architecture for vegetable disease detection is based on the latest advancements in deep learning techniques. In this project, we evaluated two types of object detection models: two-stage and one-stage object detection methods.The two-stage object detection models we tested include Faster R-CNN, Cascade Faster R-CNN, Feature Pyramid Network (FPN), and Cascade FPN. These models are considered state-of-the-art in the field of object detection. The two-stage networks generate predictions by leveraging features from both the region proposal network (RPN) and the object detector. Although these networks tend to have higher performance, they are relatively slower, making them less suitable for real-time applications such as mobile applications. To address the need for real-time applications, we employed single-stage detectors, which encompass various network architectures such as SSD, CenterNet, YOLOv3, YOLOv4, YOLOv5, and EfficientDet. These single-stage detectors offer faster inference times while still providing reliable performance for vegetable disease detection.

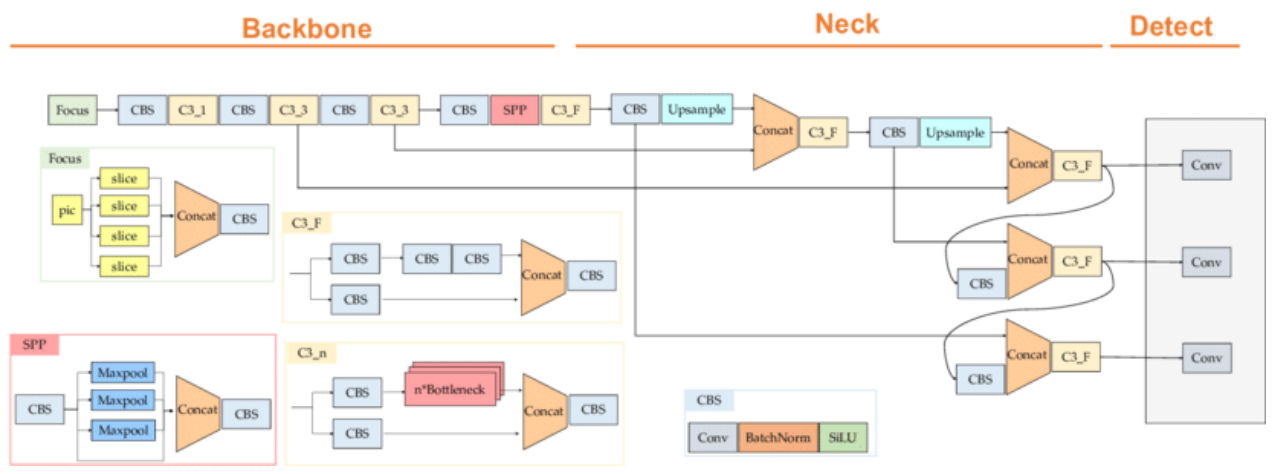

Fig. 5. YOLOv5 architecture used in this vegetable disease detection system. The detail of the architecture is available in its original paper "TPH-YOLOv5: Improved YOLOv5 Based on Transformer Prediction Head for Object Detection on Drone-captured Scenarios " available online at: https://arxiv.org/abs/2108.11539

The one-stage detector directly produces the final detection result through a single detection process. These networks are lightweight and well-suited for real-time applications. For our object detection network, we selected YOLOv5 due to its efficiency and effectiveness in various scenarios, surpassing the performance of two-stage detector networks. Figure 5 illustrates the detailed architecture of YOLOv5.

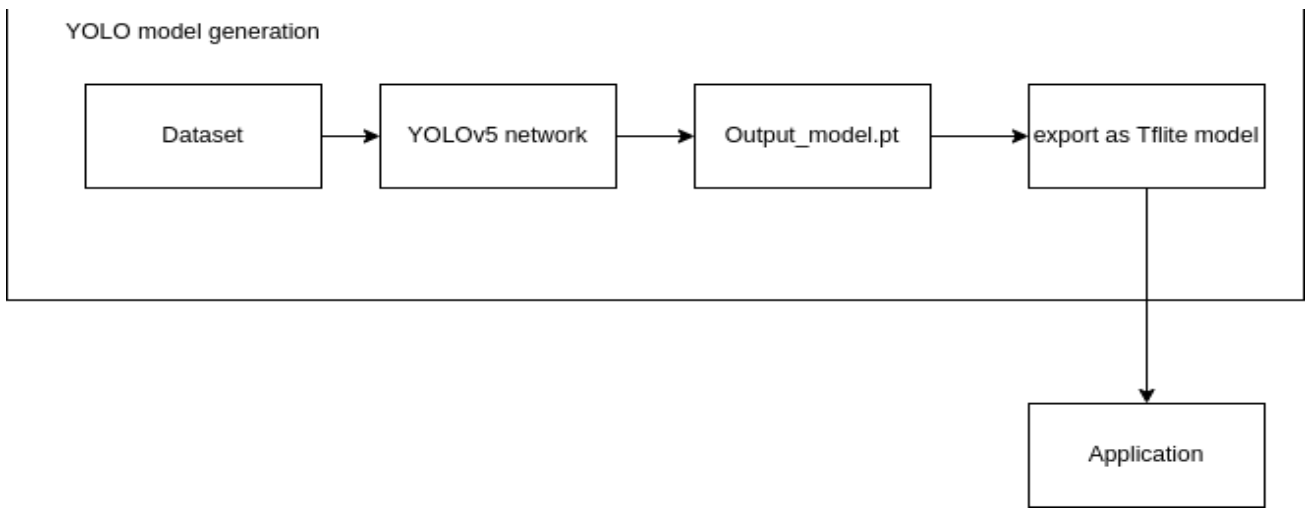

Fig. 6. Object detection model generation for mobile application

Compared to YOLOv3, YOLOv5 introduces the use of "Focus" layers. These layers perform slice operations on the input image before it enters the backbone network. This approach reduces the number of parameters to train, resulting in higher speed and preserving key information. Consequently, the width and height (W and H) information are concentrated in the channel space, amplifying them fourfold without any loss of important features. To address the issue of limited data, we employed several augmentation methods discussed in the tomato disease dataset section. These techniques helped augment and diversify the dataset, enhancing the training process. After training YOLOv5, we deployed the model in our mobile application, as depicted in Figure 6.

## B. *Mobile application*

The mobile application has been specifically designed to offer a user-friendly interface to the end-users. To store information about tomato diseases, including disease names, symptoms, and remedies, we utilized MongoDB as our database management system. The server system is connected to both the database and the user interface, as depicted in Figure 7. This architecture ensures smooth communication and seamless retrieval of information for the users of the mobile application.

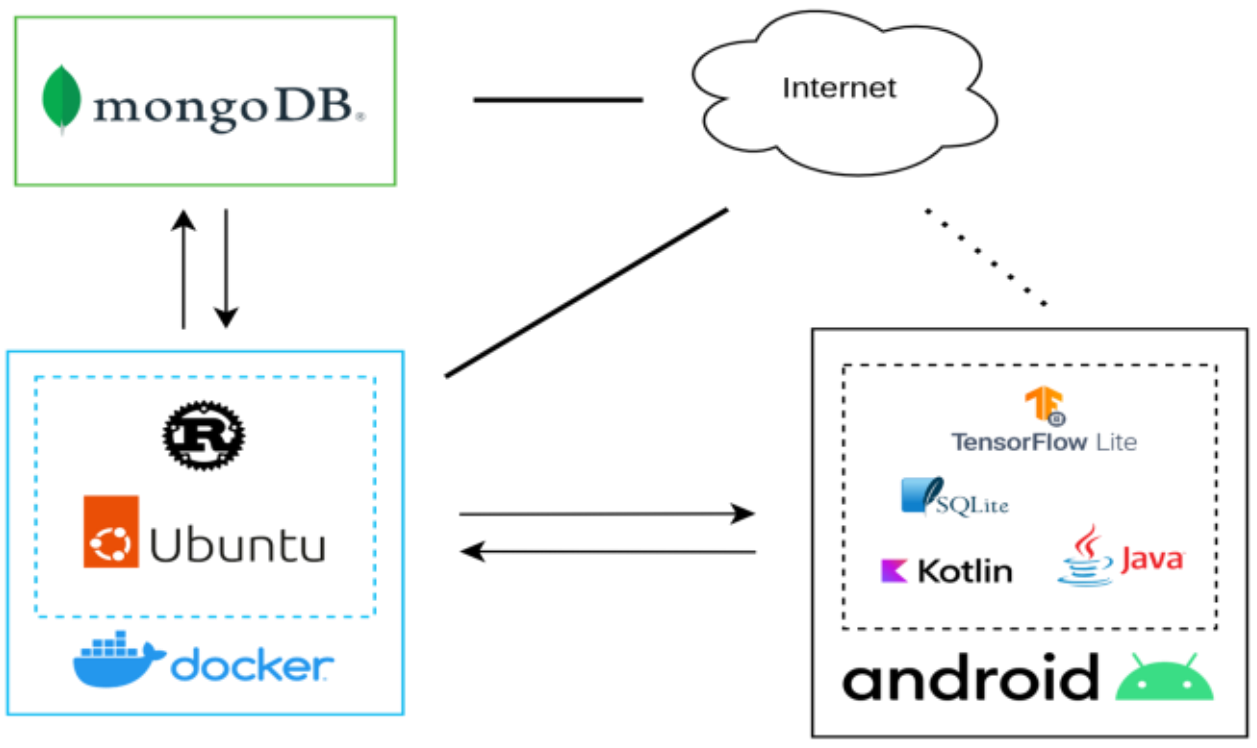

Fig. 7. Block diagram of mobile application

The detailed workflow diagram of the client architecture is depicted in Figure 8. The Android application, serving as the client system, is capable of requesting services from the server and retrieving information from the database. It allows for updates to be made to both the database and the trained neural network model as needed. This flexibility ensures that the application remains up-to-date and adaptable to changing requirements.

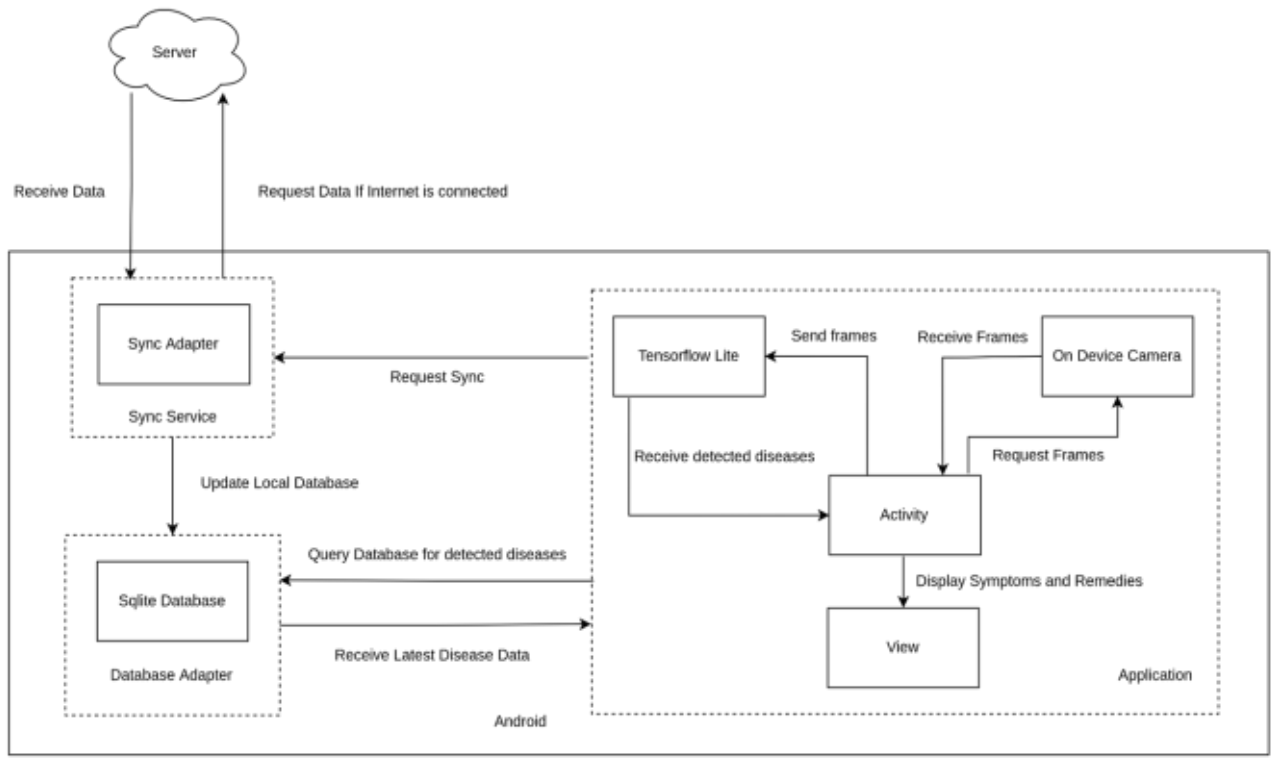

Fig. 8. Detialed diagram of client architecture

## C. *Software and hardware confuguration*

The deep neural network was trained and tested in a Python environment. The training process involved the combination of PyTorch with CUDA version 10.0 and cuDNN version 8.0. The training and inference tasks were executed on hardware consisting of an NVIDIA RTX 6000 GPU and an Intel i7-6700k CPU running at 4GHz.The mobile application was developed in a client-server architecture. The client system was built using Kotlin/Java programming languages and utilized an SQLite database. The TensorFlow Lite neural network, trained using deep learning-based object detection methods, was integrated into the client system. On the other hand, the server system was deployed on a Linux platform. The web framework was implemented using the Rust programming language, ensuring efficient and secure server-side operations. Docker technology was employed to ensure environment independence and facilitate seamless deployment.

## IV. Result and Discussion

This research focuses on the development of an artificial intelligence-based system for the detection, classification, and suggestions related to vegetable diseases. The primary aim is to assist local farmers in accurately classifying tomato diseases and providing them with a well-defined, cost-effective methodology to address this issue. To achieve this, we trained and tested both two-stage and one-stage object detection neural networks and compared their results, as outlined in Table 2. The findings indicate that the one-stage object detection method utilized in this work may be less efficient than the two-stage methods. However, it offers the advantage of being more lightweight, which is a fundamental requirement for real-time applications.

TABLE II. Comparison of two-stage and one-stage object detection methods

| Methods | Neural networks | Mean AP |
|---|---|---|
| **Two-stage object detector** | Faster RCNN with VGG-16 | 0.8306 |
| | Faster RCNN with ResNet-50 | 0.7537 |
| | Faster RCNN with ResNet-101 | 0.590 |
| | Faster RCNN with ResNet-152 | 0.6683 |
| | Faster RCNN with ResNeXt-50 | 0.711 |
| **One-stage object detector** | YOLO v5 | 0.761 |

This work utilizes the one-stage object detection model, specifically YOLOv5. The results obtained are satisfactory, as demonstrated in Figure 9 and Figure 10. The system successfully identifies the presence of grey mold disease in tomatoes at various locations on the plant, even when the ground truth labels indicate only one position. This suggests that the system has the ability to detect small disease areas appearing in multiple locations within the input image. However, Figure 10 also indicates that the system encounters some confusion during the detection of tomato diseases, leading to false positive detections. This is primarily due to the absence of a nutrex disease object in the ground truth labels of the test image. The system's performance in identifying this particular disease is limited due to the lack of training samples for nutrex disease, as illustrated in Figure 3. Therefore, with an increased number of training samples in the future, such false detections can be minimized.

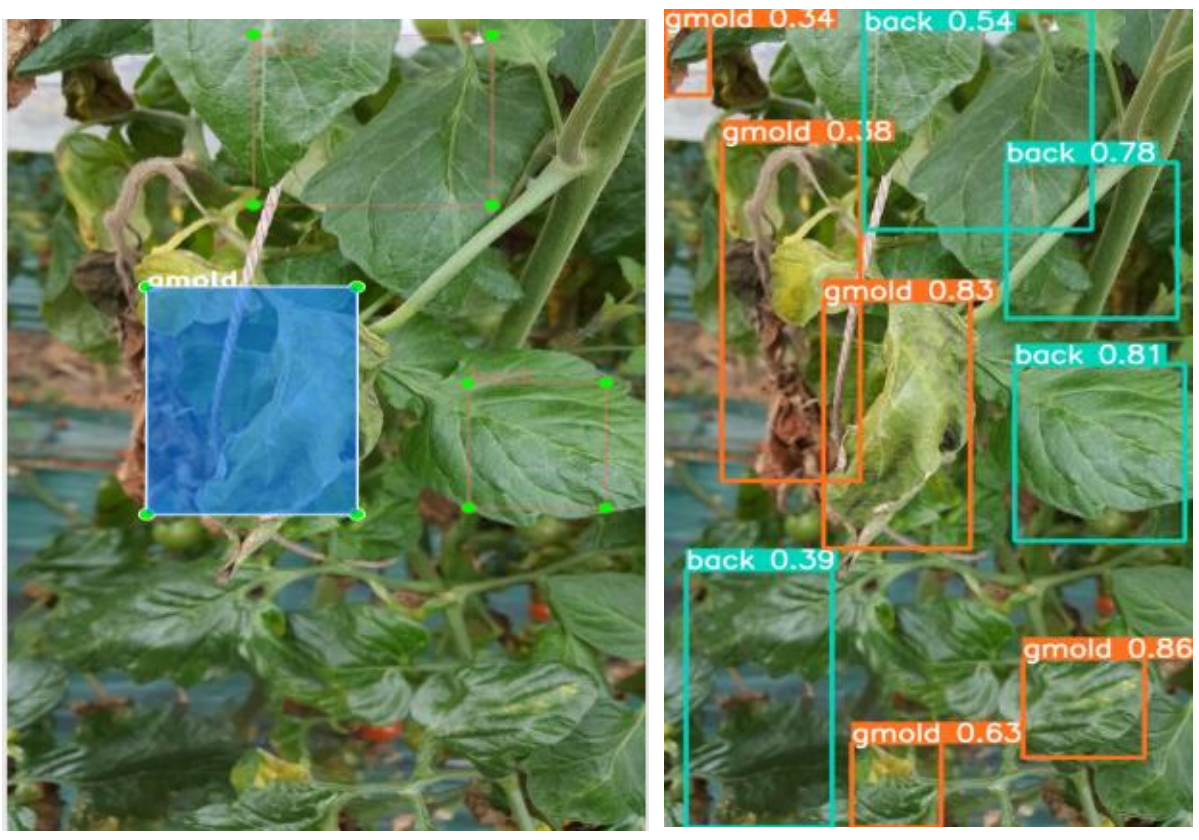

Fig. 9. The prediction results using YOLOv5 on test dataset. Left: Labelled test data, Right: Prediction from our model, back: background, gmold: grey mold disease.

The YOLOv5 model is employed for real-time vegetable disease identification. The predicted results obtained from the model are subsequently analyzed and utilized within a mobile application as part of a recommendation system. This application provides users with symptom information and remedy suggestions based on the detected disease. Figure 11 and Figure 12 showcase real-time results obtained through the mobile application for tomato disease detection and the corresponding remedy suggestion system. These figures demonstrate the practical application and functionality of the mobile application in assisting farmers with identifying tomato diseases and providing appropriate remedies.

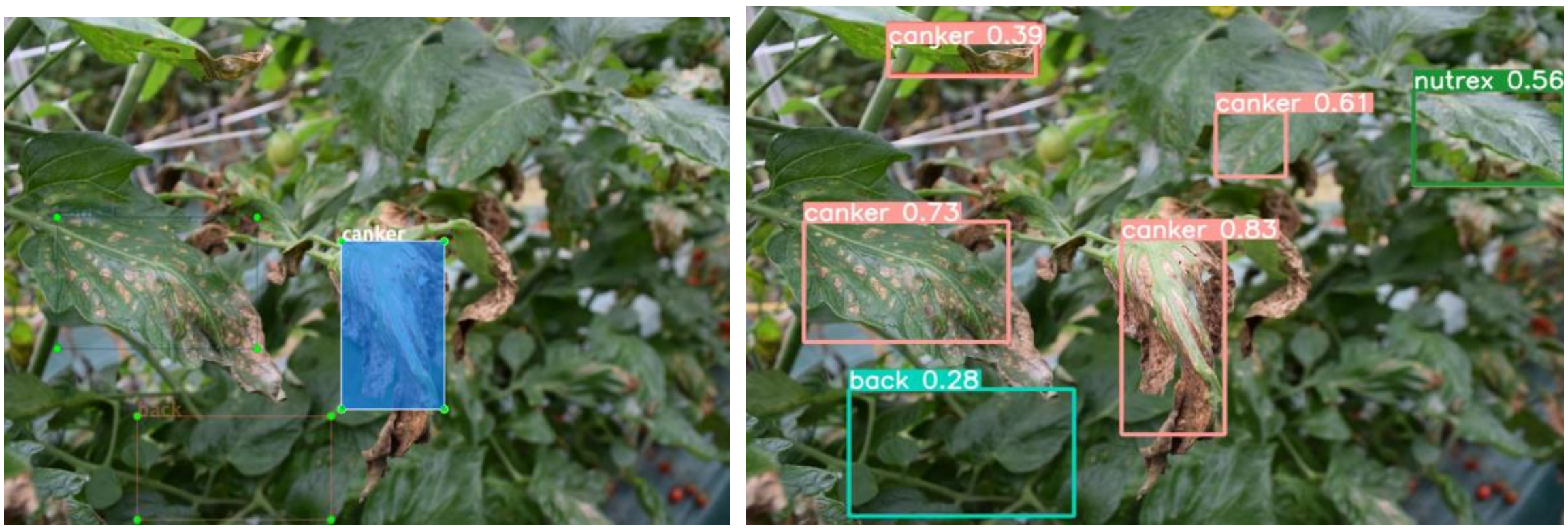

Fig. 10. The prediction results using YOLOv5 on test dataset. Left: Labelled test data, Right: Prediction from our model, back: background, canker: canker disease, nutrex: nutritional excess or deficiency

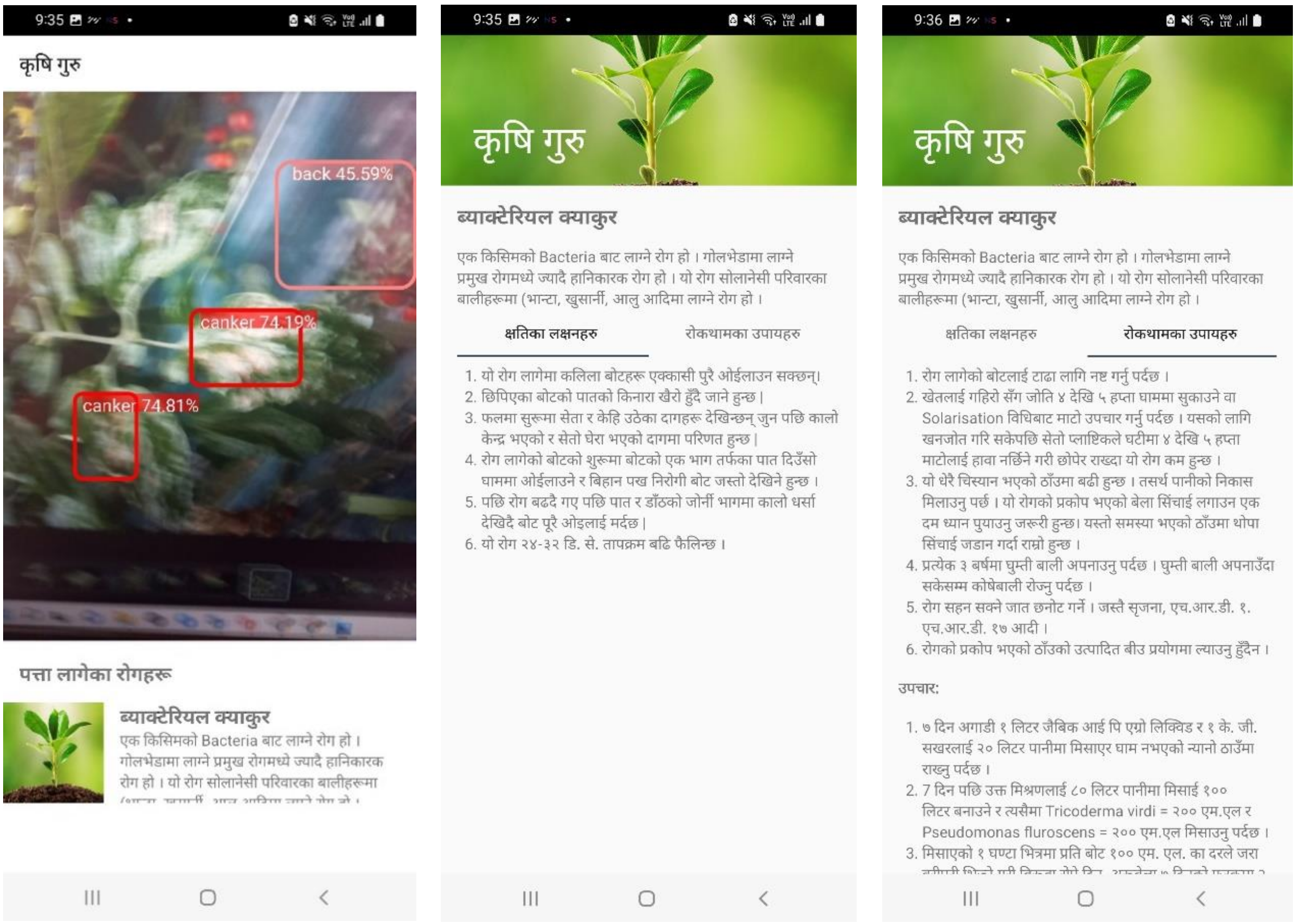

Fig. 11. Real-time tomato disease detection for canker tomato disease

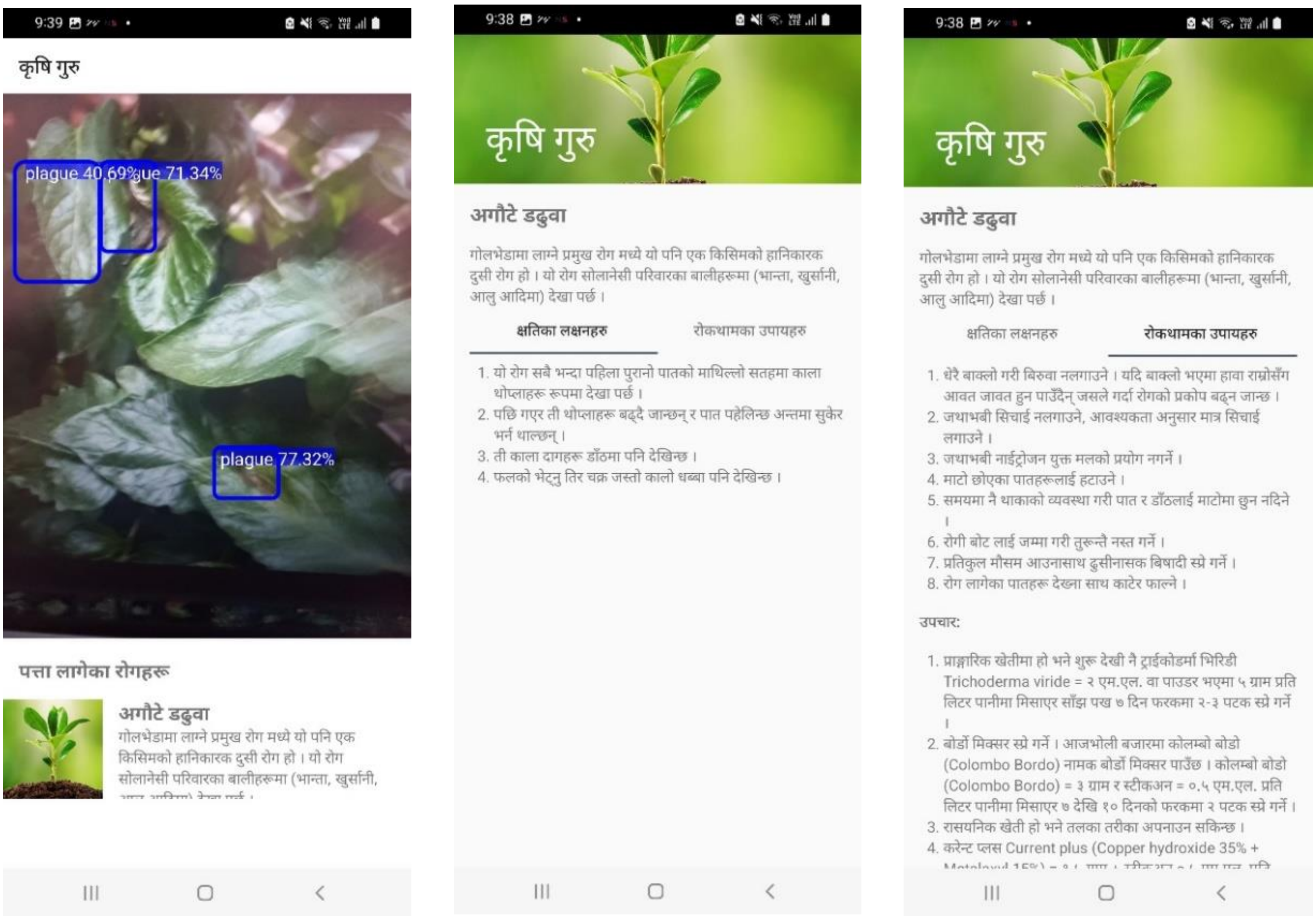

Fig. 12. Real-time tomato disease detection for plague tomato disease

## V. CONCLUSION AND FUTURE WORK

This project focuses on the development of an intelligent vegetable disease recognition system. The aim is to reduce the time and cost associated with identifying vegetable diseases, eliminating the need for constant consultation with agricultural experts. The research endeavors to contribute to quality control efforts and mitigate the misuse or overuse of hazardous pesticides. To ensure ease of use for local farmers in Nepal, the system incorporates a user-friendly interface with clear and understandable Nepali language. By providing expert suggestions and auto-guidance, the system assists local farmers in effectively addressing vegetable diseases in their farmland. The mobile application is designed to be easily accessible and free to use, making it an efficient and practical technology for local farmers to adopt.

While conducting this work, some limitations were observed, such as low confidence scores, a limited variety of diseases, and instances of false detection. However, these limitations are being actively addressed as the project remains a work in progress. The ongoing efforts involve expanding the system to accommodate multiple vegetable varieties. Valuable suggestions gathered from local farmers and agricultural experts during the final project presentation will also be considered and incorporated into future work.

## ACKNOWLEDGMENT

We would like to express our gratitude to the Kathmandu University, Community Engagement Division, Office of Vice-Chancellor for funding and supporting in research and system development of this project.